\documentclass[letterpaper, 10 pt, conference]{ieeeconf}  

\IEEEoverridecommandlockouts                              

\usepackage{xifthen}
\usepackage{xparse}
\usepackage{subdepth}
\usepackage{leftidx}
\usepackage{bm}

\newcommand{\bbm}{\begin{bmatrix}}
\newcommand{\ebm}{\end{bmatrix}}

\DeclareMathAlphabet{\mbf}{OT1}{ptm}{b}{n}
\newcommand{\mbs}[1]{{\bm{#1}}}

\newcommand{\mbsbar}[1]{{\overline{\boldsymbol{#1}}}}
\newcommand{\mbshat}[1]{{\hat{\boldsymbol{#1}}}}
\newcommand{\mbstilde}[1]{{\tilde{\boldsymbol{#1}}}}

\newcommand{\mbsdot}[1]{{\dot {\boldsymbol{#1}}}}

\newcommand{\mbfbar}[1]{{\overline{\mbf{#1}}}}
\newcommand{\mbfhat}[1]{{\hat{\mbf{#1}}}}
\newcommand{\mbftilde}[1]{{\tilde{\mbf{#1}}}}

\newcommand{\mbfdot}[1]{{\dot{\mbf{#1}}}}

\DeclareMathAlphabet{\mathbfit}{OML}{cmm}{b}{it}

\newcommand{\vel}[3]{\leftidx{_{#1}}{\mbf v}{\IfValueTF{#2}{_{#2#3\hspace{2pt}}}{}}} 
\newcommand{\veltilde}[3]{\leftidx{_{#1}}{\mbftilde v}{\IfValueTF{#2}{_{#2#3\hspace{2pt}}}{}}} 
\newcommand{\velbar}[3]{\leftidx{_{#1}}{\mbfbar v}{\IfValueTF{#2}{_{#2#3\hspace{2pt}}}{}}} 
\newcommand{\velhat}[3]{\leftidx{_{#1}}{\mbfhat v}{\IfValueTF{#2}{_{#2#3\hspace{2pt}}}{}}} 
\newcommand{\veldot}[3]{\leftidx{_{#1}}{\mbfdot v}{\IfValueTF{#2}{_{#2#3\hspace{2pt}}}{}}} 

\newcommand{\acc}[3]{\leftidx{_{#1}}{\mbf a}{\IfValueTF{#2}{_{#2#3\hspace{2pt}}}{}}} 
\newcommand{\acctilde}[3]{\leftidx{_{#1}}{\mbftilde a}{\IfValueTF{#2}{_{#2#3\hspace{2pt}}}{}}} 
\newcommand{\accbar}[3]{\leftidx{_{#1}}{\mbfbar a}{\IfValueTF{#2}{_{#2#3\hspace{2pt}}}{}}} 

\newcommand{\rotvel}[3]{\leftidx{_{#1}}{\mbs \omega}{\IfValueTF{#2}{_{#2#3\hspace{2pt}}}{}}} 
\newcommand{\rotveltilde}[3]{\leftidx{_{#1}}{\mbstilde \omega}{\IfValueTF{#2}{_{#2#3\hspace{2pt}}}{}}} 
\newcommand{\rotvelbar}[3]{\leftidx{_{#1}}{\mbsbar \omega}{\IfValueTF{#2}{_{#2#3\hspace{2pt}}}{}}} 
\newcommand{\rotvelhat}[3]{\leftidx{_{#1}}{\mbshat \omega}{\IfValueTF{#2}{_{#2#3\hspace{2pt}}}{}}} 
\newcommand{\rotveldot}[3]{\leftidx{_{#1}}{\mbsdot \omega}{\IfValueTF{#2}{_{#2#3\hspace{2pt}}}{}}} 

\usepackage{graphics} 
\usepackage{epsfig} 
\usepackage{amsmath} 
\usepackage{amssymb}  
\usepackage{amsfonts}
\usepackage[noadjust]{cite}
\usepackage{color}
\usepackage{multirow}
\usepackage{array}
\usepackage{caption}
\usepackage{booktabs}
\usepackage{makecell}
\usepackage{float}
\usepackage{flushend}

\title{\LARGE \bf
Towards the Probabilistic Fusion of Learned Priors into Standard Pipelines for 3D Reconstruction
}

\author{Tristan Laidlow$^{1}$, Jan Czarnowski$^{1}$, Andrea Nicastro$^{1}$, Ronald Clark$^{1}$ and Stefan Leutenegger$^{2}$%
\thanks{$^{1}$Dyson Robotics Laboratory at Imperial College, Imperial College London, UK}%
\thanks{$^{2}$Smart Robotics Lab, Imperial College London, UK}%
\thanks{Research presented in this paper has been supported by Dyson Technology Ltd.~and Imperial College London.  Corresponding author: Tristan Laidlow, {\tt\small t.laidlow15@imperial.ac.uk}}%
}

\begin{document}

\maketitle
\thispagestyle{empty}
\pagestyle{empty}

\begin{abstract}

The best way to combine the results of deep learning with standard 3D reconstruction pipelines remains an open problem.
While systems that pass the output of traditional multi-view stereo approaches to a network for regularisation or refinement currently seem to get the best results, it may be preferable to treat deep neural networks as separate components whose results can be probabilistically fused into geometry-based systems.
Unfortunately, the error models required to do this type of fusion are not well understood, with many different approaches being put forward.
Recently, a few systems have achieved good results by having their networks predict probability distributions rather than single values.
We propose using this approach to fuse a learned single-view depth prior into a standard 3D reconstruction system.

Our system is capable of incrementally producing dense depth maps for a set of keyframes.
We train a deep neural network to predict discrete, nonparametric probability distributions for the depth of each pixel from a single image.
We then fuse this ``probability volume'' with another probability volume based on the photometric consistency between subsequent frames and the keyframe image.
We argue that combining the probability volumes from these two sources will result in a volume that is better conditioned.
To extract depth maps from the volume, we minimise a cost function that includes a regularisation term based on network predicted surface normals and occlusion boundaries.
Through a series of experiments, we demonstrate that each of these components improves the overall performance of the system.

\end{abstract}

\section{INTRODUCTION}

\begin{figure}[t!]
  \centering
  \includegraphics[width=1\linewidth]{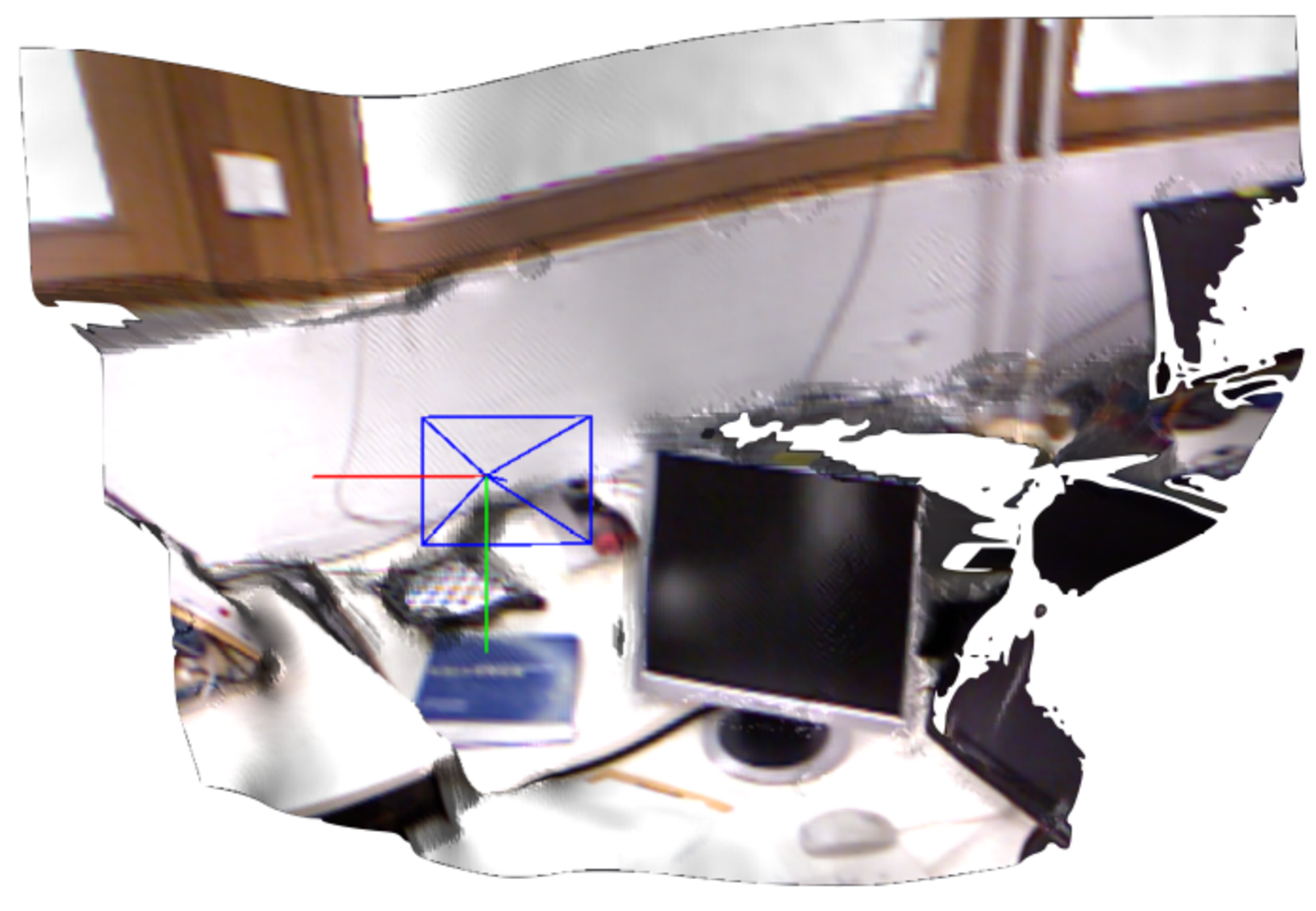}
  \caption{Fusing a single-view depth probability distribution predicted by a DNN with standard photometric error terms helps to resolve ambiguities in the photometric error due to occlusions or lack of texture. The above projected keyframe depth map was created by our system.}
  \label{fig:teaser}
  \vspace{2mm}\hrule
\end{figure}

There has been continued research interest in using structure-for-motion (SfM) and visual simultaneous localisation and mapping (SLAM) for the incremental creation of dense 3D scene geometry due to its potential applications in safe robotic navigation, augmented reality, and manipulation.
Until recently, dense monocular reconstruction systems typically worked by minimising the photometric error over several frames.
As this minimisation problem is not well-constrained due to occlusion boundaries or regions of low texture, most reconstruction systems employ regularisers based on smoothness (\cite{Newcombe:etal:ICCV2011}, \cite{Pizzoli:etal:ICRA2014}) or planar (\cite{Concha:etal:RSS2014, Concha:Civera:IROS2015, Concha:etal:ICRA2016}) assumptions.

With the continued success of deep learning in computer vision, there have been many suggestions for data-driven approaches to the monocular reconstruction problem.
Several of these approaches propose a completely end-to-end framework, predicting the scene geometry from either a single image (\cite{Eigen:etal:NIPS2014, Laina:etal:3DV2016, Godard:etal:CVPR2017, Fu:etal:CVPR2018}) or several consecutive frames (\cite{Ummenhofer:etal:CVPR2017, Zhou:etal:CVPR2017, Mahjourian:etal:CVPR2018, Zhou:etal:ECCV2018, Yao:etal:ECCV2018, Chang:Chen:CVPR2018}).
Most promising, however, are those systems that combine deep learning with standard geometric constraints (\cite{Weerasekera:etal:ICRA2017, Tateno:etal:CVPR2017, Yang:etal:ECCV2018, Bloesch:etal:CVPR2018, Wang:etal:CVPR2018, Laidlow:etal:ICRA2019, Tang:Tan:2019}).
It was shown in \cite{Facil:etal:RAL2017} that learning-based and geometry-based approaches have a complementary nature as learning-based systems tend to perform better on the interior points of objects but blur edges, whereas geometry-based systems typically do well on areas with a high image gradient but perform poorly on interior points that may lack texture.

The optimal way to combine these two approaches, however, is not clear.
The best current results seem to come from systems that take the output of traditional geometry-based systems and feed these into a deep neural network (DNN). A particularly impressive example of this type of system is DeepTAM \cite{Zhou:etal:ECCV2018}, which passes a photometric cost volume through a network to extract a depth map.

It may be desirable, however, to use learning-based systems as an additional component that is fused into the pipeline of a traditional system.
This approach would keep the probabilistic framework of the reconstruction system, a requirement for many robotic applications.
Possible benefits of such a framework include avoiding the necessity of having to perform an expensive neural network pass every time the geometric information is updated, and, as DNNs perform best on images close to the training dataset, it might be possible to switch the network component on or off or switch between different networks depending on the environment being reconstructed.
The difficulty of this approach, however, is that to probabilistically fuse the network outputs into a 3D reconstruction system, some measure of the uncertainty associated with each prediction is required.

In this paper, we propose a 3D reconstruction system that fuses together the output of a DNN with a standard photometric cost volume to create dense depth maps for a set of keyframes.
We train a network to predict a discrete, nonparametric probability distribution for the depth of each pixel over a given range from a single image.
Like \cite{Liu:etal:CVPR2019}, we refer to this collection of probability distributions for each pixel in the keyframe as a ``probability volume''.
Then, with each subsequent frame, we create a probability volume based on the photometric consistency between the current frame and the keyframe image and fuse this into the keyframe volume.
The main contribution of this paper is to demonstrate that combining the probability volumes from these two sources often results in a better conditioned probability volume.
We extract depth maps from the probability volume by optimising a cost function that includes a regularisation term based on network predicted surface normals and occlusion boundaries.
Please see Figure \ref{fig:teaser} for an example keyframe reconstruction created by our system.


\section{RELATED WORKS}

In general, uncertainty can be classified into two categories: model or epistemic uncertainty, and statistical or aleatoric uncertainty. In \cite{Gal:Ghahramani:ICML2016}, the authors suggest using a Monte Carlo dropout technique to estimate the model uncertainty of a network, but this requires multiple expensive network passes.

Like \cite{Bishop:MDN}, the authors of \cite{Kendall:Gal:NIPS2017} propose having the network predict its own aleatoric uncertainty and using a Gaussian or Laplacian likelihood as the loss function during training, which was used by \cite{Laidlow:etal:ICRA2019} for 3D reconstruction.
The problem with this approach is that it forces the network to predict a parametric and unimodal distribution.
As shown in \cite{Campbell:etal:ECCV2008}, this type of distribution may be particularly ill-suited to dense reconstruction where there is a clear need for a multi-hypothesis prediction.

One proposal has been to use a multi-headed network (\cite{Zhou:etal:ECCV2018}, \cite{Peretroukhin:etal:CVPRW2019}) with each head making a separate prediction.
From these many predictions, one can calculate the mean and covariance to use in a probabilistic fusion algorithm.
The drawbacks of this approach are that it increases the size of the network and requires a careful balancing of the relative size of the network body and heads. 

Recently, both \cite{Fu:etal:CVPR2018} and \cite{Liu:etal:CVPR2019} achieved impressive results by having their networks predict discrete, nonparametric probability distributions.
While \cite{Liu:etal:CVPR2019} uses these distributions to fuse the output with other network predictions, to the best of our knowledge, no one has used this method to fuse the predictions of networks with the output of standard reconstruction pipelines, which is what we aim to do in this paper.

\section{METHOD}

In this section, we describe our method for fusing predictions from DNNs into a standard 3D reconstruction pipeline to produce dense depth maps.

Our system represents the observed geometry as a collection of keyframe-based ``probability volumes''.
That is, instead of representing the surface as a depth map with a single depth estimate per pixel, the depth is represented with a per-pixel discrete probability distribution over a given depth range.
These probability volumes are initialised with the output of a monocular depth prediction network.
With each additional RGB image, the system computes a cost volume based on the photometric consistency.
This cost volume is then converted to a probability volume and fused into the volume of the current keyframe.
Once the number of inliers drops below a given threshold, a new keyframe is created. To propagate information from one keyframe to another, we warp the previous distribution and fuse it into the new one.

When we want to extract a depth map from the probability volume, we could take the maximum probability depth values, but in featureless regions where there is also high network uncertainty this would be susceptible to false minima and cause local inconsistencies in the prediction.
Also, as the probability distribution is discrete, taking the maximum would result in a quantisation of the final depth prediction.
To overcome these shortcomings, we first construct a smooth probability density function (PDF) from the volume using a kernel density estimation (KDE) technique.
We then minimise the negative log probability of this PDF along with a regularisation term.
While many dense systems propose using regularisers based on smoothness (\cite{Newcombe:etal:ICCV2011, Pizzoli:etal:ICRA2014}) or planar (\cite{Concha:etal:RSS2014, Concha:etal:ICRA2014, Concha:Civera:IROS2015}) assumptions, we follow the examples of \cite{Weerasekera:etal:ICRA2017} and \cite{Laidlow:etal:ICRA2019} and penalise our reconstruction for deviating from the surface normals predicted by a DNN.

\subsection{Multi-Hypothesis Monocular Depth Prediction}

Rather than predict a single depth value for each pixel, our network predicts a discrete depth probability distribution over a given range, similar to \cite{Fu:etal:CVPR2018} and \cite{Liu:etal:CVPR2019}.
Not only does this allow the network to express uncertainty about its prediction, but it also allows the network to make a multi-hypothesis depth prediction.
As discussed in \cite{Fu:etal:CVPR2018}, the prediction of the depth probability distribution can be improved by having a variable resolution over the depth range.
We choose a log-depth parameterisation, following the examples of \cite{Weerasekera:etal:ICRA2018} and \cite{Eigen:etal:NIPS2014}.
By uniformly dividing the depth range in log-space, we achieve the desired result of having higher resolution in the areas close to the camera and lower resolution farther away.

\begin{figure}
  \centering
  \includegraphics[width=1\linewidth]{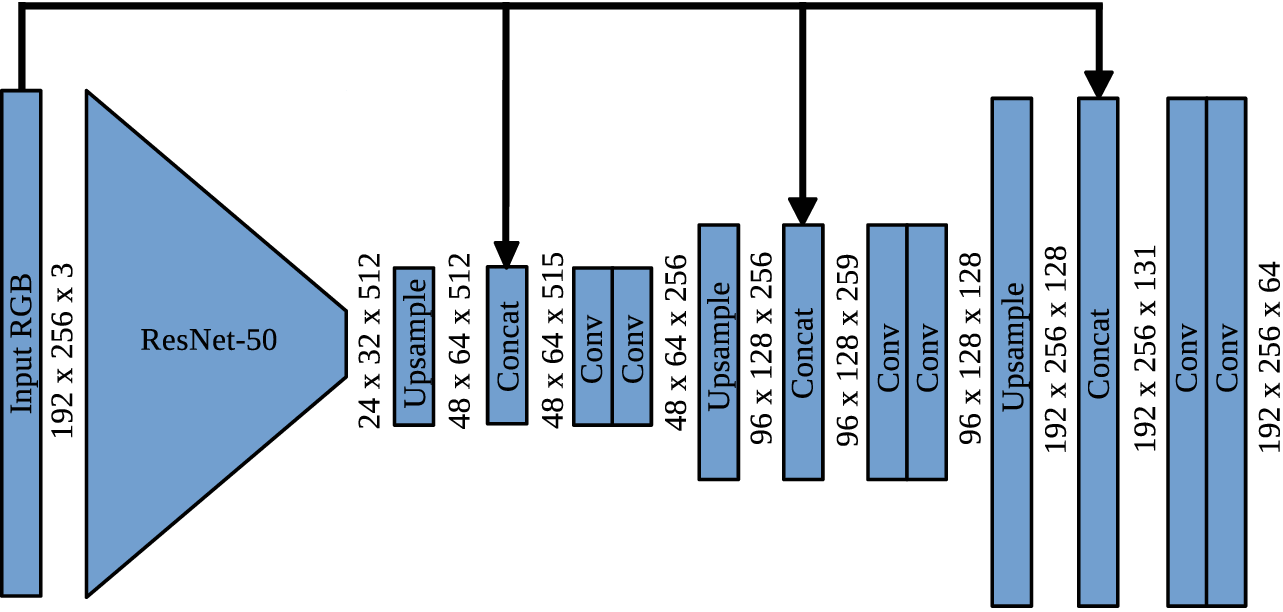}
  \caption{Our network consists of a ResNet-50 encoder with an output stride size of 8 and no global pooling layer. We then pass the output of the encoder through three upsample blocks consisting of a bilinear resize, concatenation with the input image, and then two convolutional layers to match the output resolution to the input. The probability distribution that the network outputs is discretised over 64 channels.}
  \label{fig:network}
  \vspace{2mm}\hrule
\end{figure}

For our network architecture (see Figure \ref{fig:network}), we use a ResNet-50 encoder \cite{He:etal:CVPR2016} followed by three upsample blocks, each consisting of a bilinear upsampling layer, a concatenation with the input image, and then two convolutional layers to bring the output back up to the input resolution.
All inputs and outputs have a resolution of 256$\times$192.

As we are having the network predict a discrete distribution rather than a depth map, we cannot use a standard loss function based on the sum of squared errors.
A cross-correlation loss would not be ideal either, as we would like to penalise the network less for predicting high probabilities in incorrect bins that are close to the true bin than in bins farther away.
Instead, we choose to use the ordinal loss function proposed in \cite{Fu:etal:CVPR2018}:
\begin{equation}
\begin{split}
    \mathcal{L}(\mbs \theta) ={} & -\sum_i {} \Biggl[ {} \sum_{k = 0}^{k_i^*} \log(p_{\mbs \theta,i}(k_i^* \geq k)) \\
    &+ \sum_{k = k_i^* + 1}^{K-1} \log(1 - p_{\mbs \theta,i}(k_i^* \geq k)) \Biggr],
\end{split}
\end{equation}
\noindent
where
\begin{equation}
    p_{\mbs \theta,i}(k_i^* \geq k) = \sum_{j = k}^{K-1} p_{\mbs \theta,i}(k_i^* = j),
\end{equation}
\noindent
$\mbs \theta$ is the set of network weights, $K$ is the number of bins over which the depth range is discretised, $k_i^*$ is the index of the bin containing the ground truth depth for pixel $i$, and $p_{\mbs \theta,i}(k_i^* = j)$ is the network prediction of the probability that the ground truth depth is in bin $j$.

Like \cite{Liu:etal:CVPR2019}, we train our network on the ScanNet RGB-D dataset \cite{Dai:etal:CVPR2017}.
No fine-tuning was done on our evaluation dataset, the TUM RGB-D dataset \cite{Sturm:etal:IROS2012}.
We set the depth range to be between 10cm and 12m and group the log-depth values uniformly into 64 bins.

Each keyframe created by our system is initialised with this network output.

\subsection{Fusion with Photometric Error Terms}

For each additional reference frame, we construct a DTAM-style cost volume \cite{Newcombe:etal:ICCV2011}.
First, we normalise both the keyframe and reference frame images by subtracting their means and dividing by their standard deviations.
We then calculate the photometric error by warping the normalised keyframe image into the reference frame for each depth value in the cost volume and taking the sum of squared differences on 3$\times$3 patches.
To simplify the later fusion, we use the midpoint of each of the depth bins used for the network prediction as the depth values in the cost volume.
Poses are obtained from an oracle, such as a separate tracking system like ORB-SLAM2 \cite{Mur-Artal:etal:TRO2017}.

To convert to a probability volume, we separately scale the negative of the squared photometric error for each pixel such that it sums to one over the ray.
We then fuse this new probability volume, $p_{\text{RF}}$, into the current keyframe volume, $p_{\text{KF}}$:
\begin{equation}
    p_i(k_i^* = k) = p_{\text{KF},i}(k_i^* = k) p_{\text{RF},i}(k_i^* = k),
\end{equation}
\noindent
for each pixel $i$ and depth $k$, which is then scaled to sum to one over the ray.

\subsection{Kernel Density Estimation}

To avoid a quantisation of the final depth prediction and to have a smooth function to use in the optimisation step, we construct a PDF for the depth of each pixel using a KDE technique with Gaussian basis functions:
\begin{equation}
    f_i(d) = \sum_{k = 0}^{K-1} p_i(k_i^* = k) \phi\left(d(k), \sigma\right)
\end{equation}
\noindent
where $\phi\left(\mu, \sigma\right)$ is the probability density of the Gaussian distribution with mean $\mu$ and standard deviation $\sigma$, $d(k)$ is the depth value at the midpoint of bin $k$, and $\sigma$ is a constant smoothing parameter across all pixels and depth values.
The value of $\sigma$ is a hyperparameter that needs to be tuned empirically; we found that $\sigma = 0.1$ works well in our setting.

\begin{figure}
  \centering
  \includegraphics[width=1\linewidth]{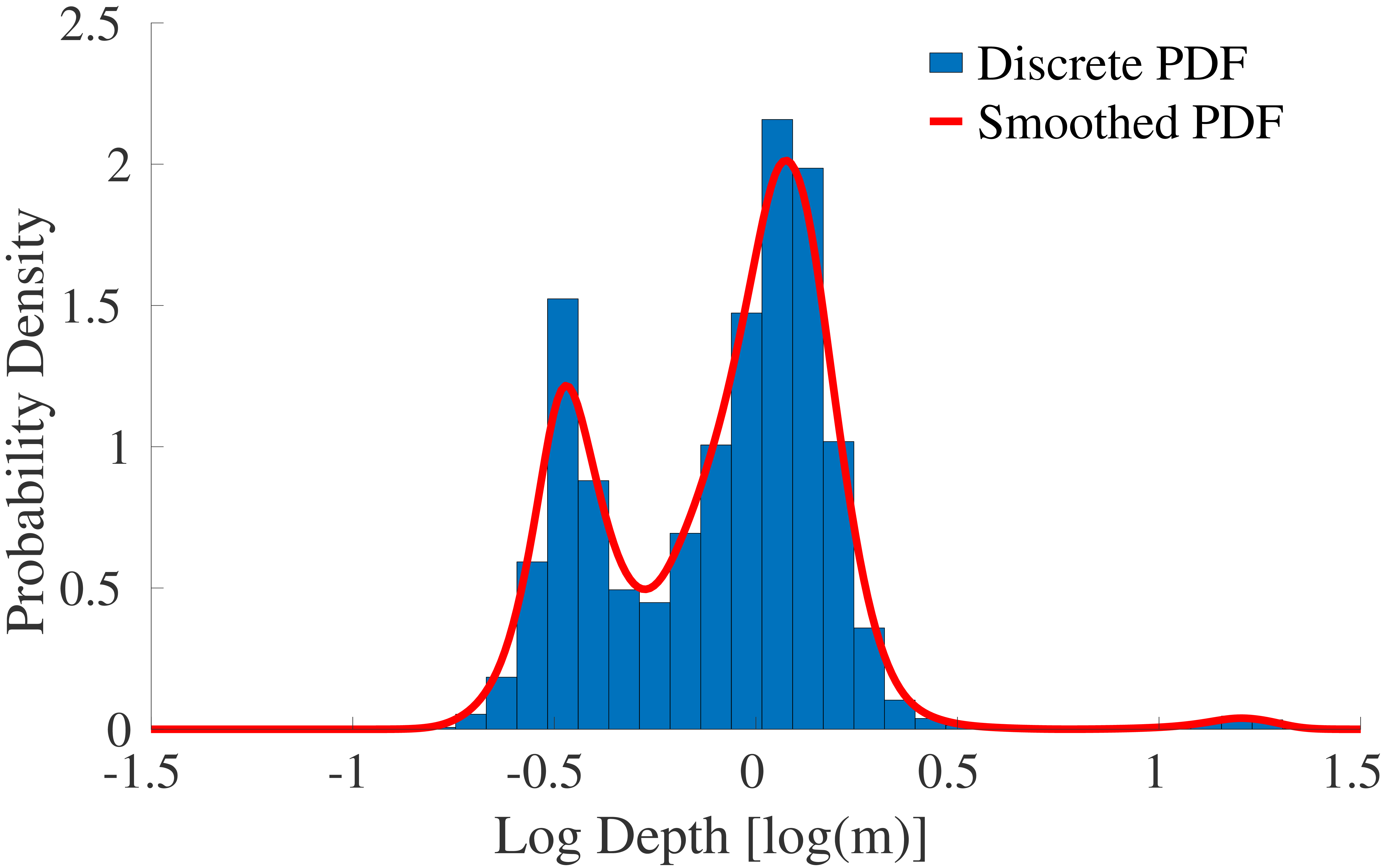}
  \caption{Our fusion algorithm produces a discrete probability distribution for each pixel in the keyframe. To reduce discretisation errors and to have a continuous cost function for the optimiser, we convert the probability values along each ray into a smooth probability density function using a kernel density estimation technique.}
  \label{fig:smoothing}
  \vspace{2mm}\hrule
\end{figure}

An example of a discrete PDF produced by our system and the smoothed result after applying the KDE technique is shown in Figure \ref{fig:smoothing}.

\subsection{Regularisation}

\begin{figure}
  \centering
  \includegraphics[width=1\linewidth]{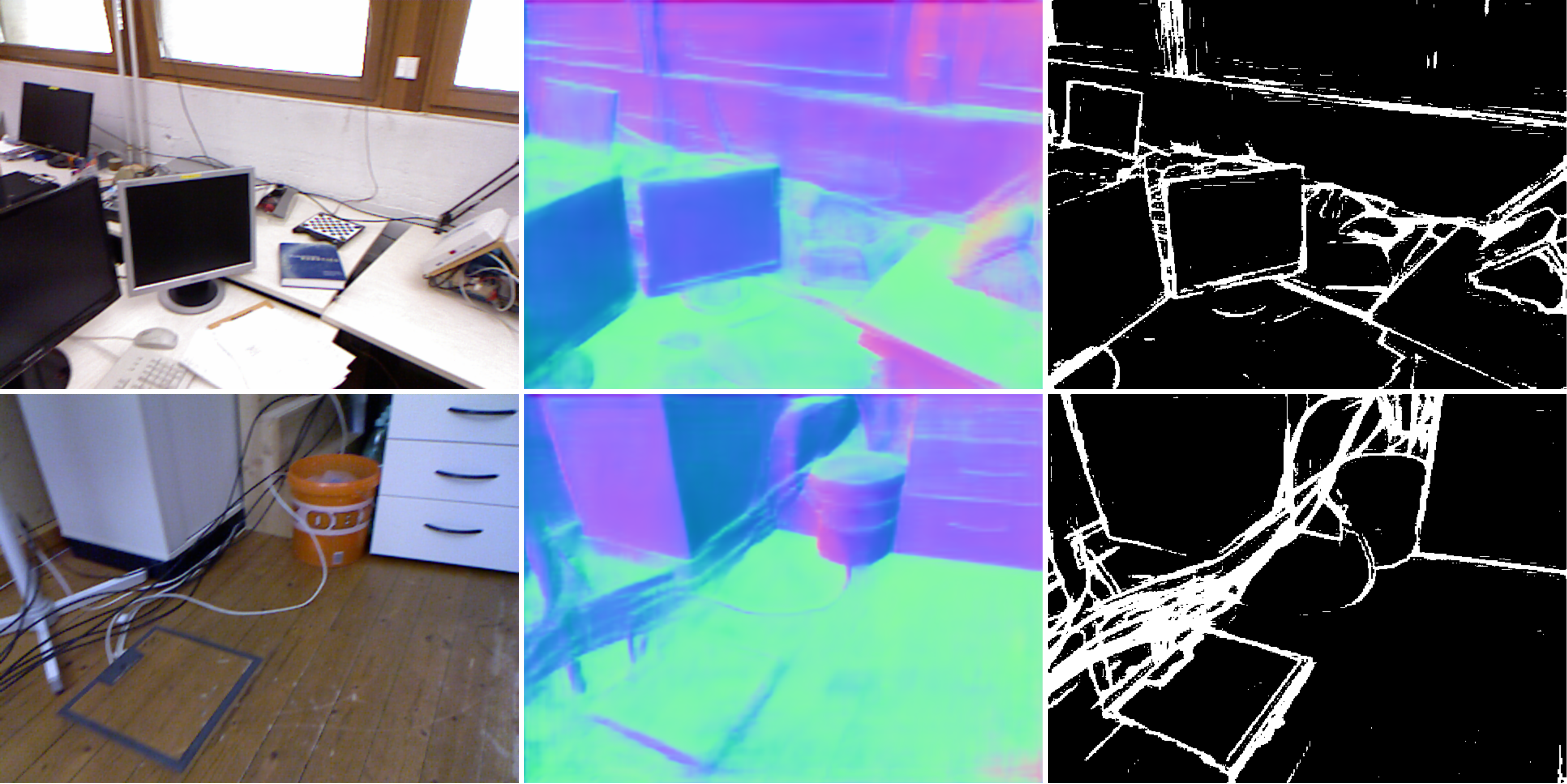}
  \caption{To regularise our depth estimate, we use the surface normals and occlusion boundaries predicted by SharpNet \cite{Ramamonjisoa:Lepetit:ICCVW2019}. Some examples of the predictions made by SharpNet on the TUM RGB-D dataset \cite{Sturm:etal:IROS2012} are shown above. From left to right: input RGB images, predicted normals, predicted occlusion boundaries with a probability greater than 0.4.}
  \label{fig:sharpnet}
  \vspace{2mm}\hrule
\end{figure}

Although the fused probability volume will have more local consistency than using the photoconsistency terms alone, the result can still be improved by adding a regularisation term to the optimisation used to extract the depth map.
While most dense reconstruction systems base their regularisers on smoothness or planar assumptions, we propose using the surface normals predicted by a DNN as was done in both \cite{Weerasekera:etal:ICRA2017} and \cite{Laidlow:etal:ICRA2019} as this may allow for better preservation of fine-grained local geometry.
To predict the surface normals from the keyframe image, we use the state-of-the-art network SharpNet \cite{Ramamonjisoa:Lepetit:ICCVW2019}.
As we determine the local surface orientation of our depth estimation from neighbouring pixels and we do not wish incur high costs at depth discontinuities, we mask the regularisation term at occlusion boundaries, which are also predicted by SharpNet.
Since SharpNet actually predicts a probability of each pixel belonging to an occlusion boundary, we include all pixels with a probability higher than 0.4 in the mask.
Example predictions of surface normals and occlusion boundaries made by SharpNet on the TUM RGB-D dataset \cite{Sturm:etal:IROS2012} are shown in Figure \ref{fig:sharpnet}.

\subsection{Optimisation}

To extract a depth map from the probability volume, we minimise a cost function consisting of two terms:
\begin{equation}
    c(\mbf{d}) = c_f(\mbf{d}) + \lambda c_{\mbf{\hat n}}(\mbf{d}),
\end{equation}
\noindent
where $\mbf{d}$ is the set of depth values to be estimated, and $\lambda$ is a hyperparameter used to adjust the strength of the regularisation term. Empirically, we found $\lambda = 1.0 \cdot 10^7$ to work well.

The first term, $c_f$, imposes a unary constraint on each of the pixels:
\begin{equation}
    c_{f}(\mbf{d}) = -\sum_i \log \left( f_i(d_i) \right)
\end{equation}
\noindent
where $f_i(d_i)$ is the smoothed PDF of pixel $i$ evaluated at depth $d_i$.

The second term, $c_{\mbf{\hat n}}$, is a regularisation term that combines two pairwise constraints:
\begin{equation}
\begin{split}
    c_{\mbf{\hat n}}(\mbf{d}) = \sum_i & b_i \left( \langle \mbf{\hat n}_i, d_i \mbf{K}^{-1} \mbf{\tilde u}_i - d_{i+1} \mbf{K}^{-1} \mbf{\tilde u}_{i+1} \rangle \right)^2 \\
    &+ b_i \left( \langle \mbf{\hat n}_i, d_i \mbf{K}^{-1} \mbf{\tilde u}_i - d_{i+\text{W}} \mbf{K}^{-1} \mbf{\tilde u}_{i+\text{W}} \rangle \right)^2
\end{split}
\end{equation}
\noindent
where $b_i \in \{0, 1\}$ is the value of the occlusion boundary mask for pixel $i$, $\langle \cdot, \cdot \rangle$ is the dot product operator, $\mbf{\hat n}_i$ is the normal vector predicted by SharpNet, $\mbf{K}$ is the camera intrinsics matrix, $\mbf{\tilde u}_i$ is the homogeneous pixel coordinates for pixel $i$, and $\text{W}$ is the width of the image in pixels.

We minimise the cost function by applying a maximum of 100 iterations of gradient descent with a step size of 0.2, and initialise the optimisation with the maximum probability depth values from the fused probability volume.
As the focus of this paper is on the benefits of fusing learning-based and geometry-based approaches, the system was not implemented to achieve real time results.
Currently, the forward pass of the network is not a major bottleneck (it takes approximately 53ms), but the process of going from a fused probability volume through the smoothing and optimisation to an extracted depth map can take up to 1.2s, depending on how many iterations are required before the stopping criterion is met.
This could be improved significantly by using Newton's method or the primal-dual algorithm instead of gradient descent; however, we leave this for future work.

\subsection{Keyframe Warping}

To avoid throwing away information on the creation of each new keyframe, we warp the probability volume of the current keyframe and use it to initialise the new one.
As the probability volume is a distribution over the depth values of a pixel, however, warping the probability volume is not trivial.
To do this, we propose using a discrete variation of the method described in \cite{Loop::etal::3DV2016}, where we first convert the depth probability distribution to an occupancy-based probability volume, where for each depth bin along the ray there is a probability that the associated point in space is occupied. 
We then warp this occupancy grid into the new frame and convert back to a depth probability distribution.

We start by defining the probability that the voxel $S_{k,i}$ (corresponding to depth bin $k$ along the ray of pixel $i$) is occupied, conditioned on the depth belonging to bin $j$:
\begin{equation}
    p(S_{k,i} = 1 \lvert k_i^* = j) =
    \begin{cases}
        0           & \text{if}\ k < j \\
        1           & \text{if}\ k = j \\
        \frac{1}{2} & \text{if}\ k > j
    \end{cases}
\end{equation}

To convert a depth probability distribution into an occupancy probability, we marginalise out the conditional:
\begin{equation}
    p(S_{k,i} = 1) = \sum_{k=0}^{K-1} p_i(k_i^* = k) p(S_{k,i} = 1 \lvert k_i^* = k)
\end{equation}
\noindent
where $p_i(k_i^* = k)$ is the value of the depth probability volume in bin $k$ for pixel $i$.

As the occupancy grid represents probabilities for locations in 3D space, we can directly warp this into the new keyframe, filling in any unknown values with a default occupancy probability.
In our work, we use a default probability of 0.01.

After warping, the occupancy grid can be converted back into a depth probability distribution:
\begin{equation}
    p_i(k_i^* = k) = \prod_{j < k} \left[1 - p(S_{j,i} = 1)\right] p(S_{k,i} = 1),
\end{equation}
\noindent
and scaled so that the distribution sums to one along the ray.

\section{EXPERIMENTAL RESULTS}

We evaluate our system on the Freiburg 1 sequences of the TUM RGB-D dataset \cite{Sturm:etal:IROS2012}.
Please note that only the RGB images are processed by our system and the depth channel is just used as a ``ground truth'' with which to validate our results against.

\subsection{Qualitative Results}

\begin{figure*}
  \centering
  \includegraphics[width=1\linewidth]{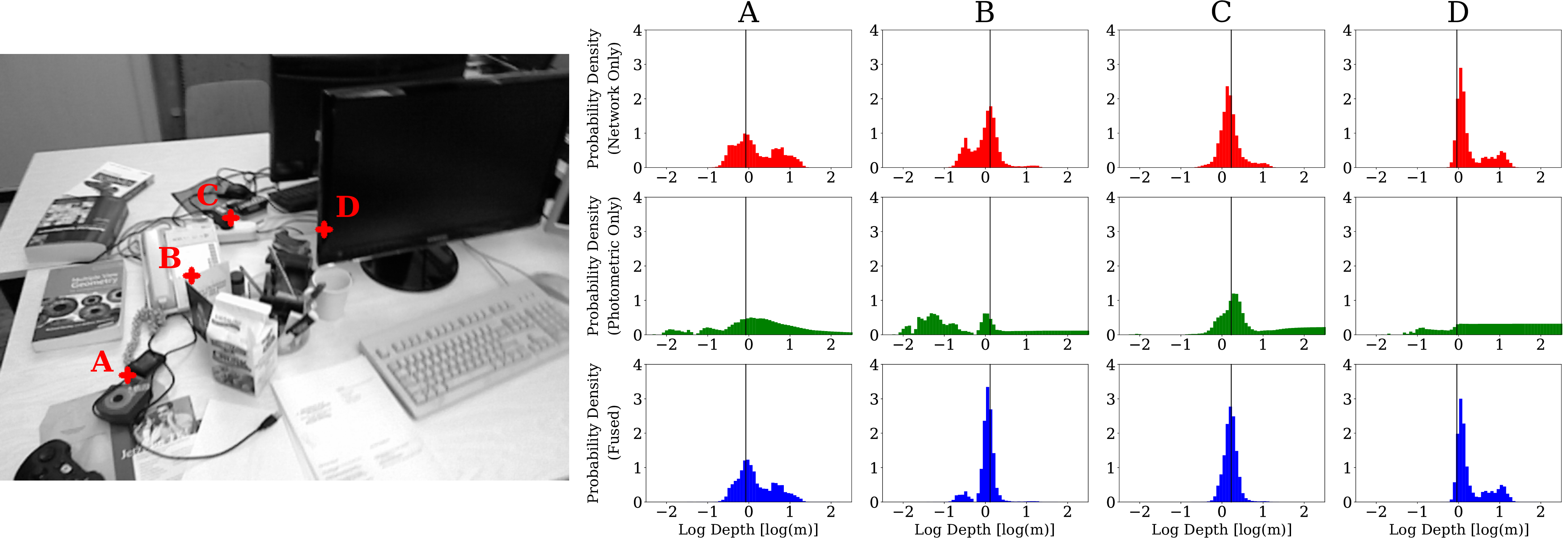}
  \caption{This figure shows a grid of probability densities for a sample of four pixels from a keyframe (left). The first row, in red, shows the probability densities predicted by the network. The second row, in green, shows the probability densities estimated from the photometric error after the addition of 25 reference frames. The final row, in blue, shows the fused probability densities that results from our algorithm. Note that both the network and the photometric error are capable of producing multiple peaks. In some cases (such as pixel C), both the network and the photometric methods produce good estimates. In others (such as pixel A), both the network and photometric error are relatively uncertain, but together produce a strong peak. In pixels B and D, the network helps resolve ambiguous photometric peaks from either a repetition or lack of texture. The vertical black bars show the location of the ground truth depth.}
  \label{fig:fusion}
  \vspace{2mm}\hrule
\end{figure*}

Figure \ref{fig:fusion} shows the various PDFs for a sample of four pixels taken from a keyframe in the TUM RGB-D sequence \textit{fr1/desk}.
The PDFs in the first row are those predicted by the DNN. Note that the network is able to make multi-hypothesis predictions and can have varying degrees of certainty.
The PDFs in the second row are those that result from the photometric cost volume.
For some of the pixels (such as pixels A and C), the photometric error results in a clear peak.
This situation is most often found on corners and edges in the image where there are large intensity gradients.
For pixels in textureless regions or on occlusion boundaries or areas with repeating patterns, the photometric PDF may have many peaks (such as pixel B) or no peak at all (such as Pixel D).
The final row of the figure shows the fused PDF for each of the pixels.
By fusing the two PDFs together, uncertainty can be reduced and ambiguous photometric data can be resolved.

\begin{figure*}
  \centering
  \includegraphics[width=1\linewidth]{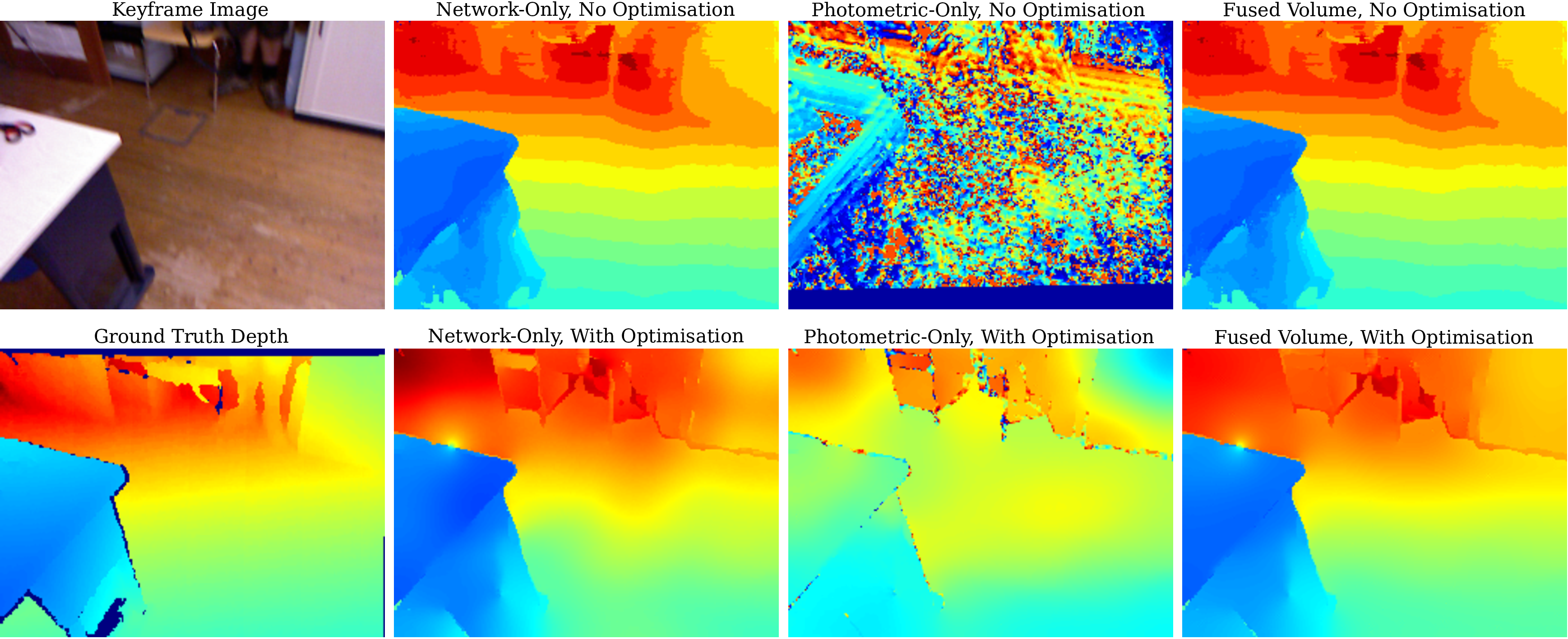}
  \caption{Qualitative results from an example keyframe and 6 additional reference frames in the TUM RGB-D \textit{fr1/360} sequence. The top left image is the keyframe image, and the bottom left is the ground truth depth. The remaining images on the top row are the depth estimates obtained by taking the maximum probability depth from each corresponding probability volume. The bands of colour show the quantisation that results from using this method. The remaining images in the bottom row are the depth estimates that result after performing the optimisation step. Note that the photometric error is only capable of estimating the depth at pixels with a high image gradient (the repeated edges are the result of pose error). While using only the network prediction results in a good reconstruction, the best reconstruction is obtained by fusing the network and photometric volumes together.}
  \label{fig:qualex}
  \vspace{2mm}\hrule
\end{figure*}

An example reconstruction for a single keyframe with various ablations is shown in Figure \ref{fig:qualex}.

\subsection{Quantitative Evaluation}

We demonstrate the value of fusion on the reconstruction pipeline by comparing the performance of the system on each of the Freiburg 1 TUM RGB-D sequences under three different scenarios: using only the network probability volume, using only the photometric probability volume, and using the fused probability volume.
To isolate the performance of our reconstruction system, we use the ground truth poses provided in the dataset.
We evaluate the performance using three metrics defined in \cite{Eigen:etal:NIPS2014}: the absolute relative difference (L1-rel), the squared relative difference (L2-rel) and the root mean squared error (RMSE).
Note that since the photometric probability volume has extremely noisy results on textureless surfaces, we found that the results were improved by initialising the optimisation with the expected value of the depth from the probability volume rather than the highest probability depth.

The results are presented in Table \ref{tab:fusion_ablation}.
In six of the sequences, the best result is achieved by fusing together the network and photometric probability volumes.
While there is a large performance gain in using the network over the photometric probability volume, the best outcome is achieved by fusing the two together.
In one of the sequences (\textit{fr1/floor}), the best result is achieved by using only the photometric probability volume.
For this entire sequence, the camera is aimed at a bare wooden floor, and, being well outside the training distribution, the network produces particularly bad priors.
In the remaining two sequences, the best result is achieved using only the network probability volume.
In one of these sequences (\textit{fr1/rpy}), the camera motion is purely rotational and the photoconsistency-based subsystem is not able to produce meaningful depth estimates.


\begin{table}[H]
\centering
\def\arraystretch{0.9}
\begin{tabular}{l l c c c}
    \toprule
    \textbf{Sequence} & \textbf{System} & \textbf{L1-rel} & \textbf{L2-rel} & \textbf{RMSE} \\
    \midrule
    \multirow{4}{*}{fr1/360}   & Network-Only     &         0.193  &         0.147  & \textbf{0.555} \\
                               & Photometric-Only &         0.633  &         1.008  &         1.514  \\
                               & Fused            & \textbf{0.191} & \textbf{0.143} & \textbf{0.555} \\
                               \cmidrule(l){2-5}
                               & DeepTAM*          &         0.194  &         0.174  &         0.559  \\
    \midrule
    \multirow{4}{*}{fr1/desk}  & Network-Only     &         0.295  &         0.201  &         0.447  \\
                               & Photometric-Only &         0.541  &         0.503  &         0.859  \\
                               & Fused            & \textbf{0.278} & \textbf{0.177} & \textbf{0.427} \\
                               \cmidrule(l){2-5}
                               & DeepTAM*          &         0.089  &         0.031  &         0.213  \\
    \midrule
    \multirow{4}{*}{fr1/desk2} & Network-Only     &         0.237  &         0.139  & \textbf{0.423} \\
                               & Photometric-Only &         0.522  &         0.494  &         0.890  \\
                               & Fused            & \textbf{0.236} & \textbf{0.138} &         0.424  \\
                               \cmidrule(l){2-5}
                               & DeepTAM*          &         0.111  &         0.049  &         0.270  \\
    \midrule
    \multirow{4}{*}{fr1/floor} & Network-Only     &         0.806  &         0.727  &         0.821  \\
                               & Photometric-Only & \textbf{0.488} & \textbf{0.303} & \textbf{0.562} \\
                               & Fused            &         0.785  &         0.691  &         0.796  \\
                               \cmidrule(l){2-5}
                               & DeepTAM*          &         0.131  &         0.034  &         0.156  \\
    \midrule
    \multirow{4}{*}{fr1/plant} & Network-Only     &         0.426  &         0.502  & \textbf{0.816} \\
                               & Photometric-Only &         0.726  &         1.422  &         1.983  \\
                               & Fused            & \textbf{0.416} & \textbf{0.485} &         0.833  \\
                               \cmidrule(l){2-5}
                               & DeepTAM*          &         0.167  &         0.143  &         0.602  \\
    \midrule
    \multirow{4}{*}{fr1/room}  & Network-Only     &         0.231  &         0.155  &         0.493  \\
                               & Photometric-Only &         0.605  &         0.762  &         1.187  \\
                               & Fused            & \textbf{0.226} & \textbf{0.147} & \textbf{0.488} \\
                               \cmidrule(l){2-5}
                               & DeepTAM*          &         0.132  &         0.079  &         0.367  \\
    \midrule
    \multirow{4}{*}{fr1/rpy}   & Network-Only     & \textbf{0.242} & \textbf{0.199} & \textbf{0.577} \\
                               & Photometric-Only &         0.514  &         0.577  &         1.047  \\
                               & Fused            &         0.255  &         0.212  &         0.614  \\
                               \cmidrule(l){2-5}
                               & DeepTAM*          &         0.154  &         0.101  &         0.427  \\
    \midrule
    \multirow{4}{*}{fr1/teddy} & Network-Only     & \textbf{0.294} & \textbf{0.271} & \textbf{0.773} \\
                               & Photometric-Only &         0.748  &         1.569  &         2.108  \\
                               & Fused            &         0.297  &         0.277  &         0.792  \\
                               \cmidrule(l){2-5}
                               & DeepTAM*          &         0.173  &         0.157  &         0.626  \\
    \midrule
    \multirow{4}{*}{fr1/xyz}   & Network-Only     &         0.241  &         0.162  &         0.432  \\
                               & Photometric-Only &         0.517  &         0.403  &         0.764  \\
                               & Fused            & \textbf{0.225} & \textbf{0.137} & \textbf{0.401} \\
                               \cmidrule(l){2-5}
                               & DeepTAM*          &         0.065  &         0.017  &         0.164  \\
    \bottomrule
\end{tabular}
\caption{Comparison of reconstruction errors on Freiburg 1 TUM RGB-D \cite{Sturm:etal:IROS2012} sequences showing the relative performance of using only the network-predicted probability volume, only the photometric probability volume, and the fused probability volume. *Despite more accurate
results, DeepTAM does not maintain a probabilistic formulation.}
\label{tab:fusion_ablation}
\vspace{2mm}\hrule
\end{table}

\begin{table}[h]
\centering
\def\arraystretch{0.9}
\begin{tabular}{l l c c c}
    \toprule
    \textbf{Sequence} & \textbf{System} & \textbf{L1-rel} & \textbf{L2-rel} & \textbf{RMSE} \\
    \midrule
    \multirow{4}{*}{fr1/360}   & No Optimisation      &         0.210  &         0.184  &         0.608  \\
                               & Smoothing-Only       &         0.207  &         0.179  &         0.601  \\
                               & Total Variation      &         0.194  &         0.152  &         0.565  \\
                               & Normals + Occlusions & \textbf{0.191} & \textbf{0.143} & \textbf{0.555} \\
    \midrule
    \multirow{4}{*}{fr1/desk}  & No Optimisation      &         0.324  &         0.292  &         0.537  \\
                               & Smoothing-Only       &         0.323  &         0.289  &         0.533  \\
                               & Total Variation      &         0.296  &         0.226  &         0.470  \\
                               & Normals + Occlusions & \textbf{0.278} & \textbf{0.177} & \textbf{0.427} \\
    \midrule
    \multirow{4}{*}{fr1/desk2} & No Optimisation      &         0.283  &         0.240  &         0.537  \\
                               & Smoothing-Only       &         0.280  &         0.235  &         0.532  \\
                               & Total Variation      &         0.254  &         0.181  &         0.470  \\
                               & Normals + Occlusions & \textbf{0.236} & \textbf{0.138} & \textbf{0.424} \\
    \midrule
    \multirow{4}{*}{fr1/floor} & No Optimisation      &         0.806  &         0.772  &         0.861  \\
                               & Smoothing-Only       &         0.807  &         0.771  &         0.860  \\
                               & Total Variation      &         0.801  &         0.738  &         0.836  \\
                               & Normals + Occlusions & \textbf{0.785} & \textbf{0.691} & \textbf{0.796} \\
    \midrule
    \multirow{4}{*}{fr1/plant} & No Optimisation      &         0.436  &         0.558  &         0.863  \\
                               & Smoothing-Only       &         0.435  &         0.555  &         0.857  \\
                               & Total Variation      &         0.425  &         0.520  & \textbf{0.830} \\
                               & Normals + Occlusions & \textbf{0.416} & \textbf{0.485} &         0.833 \\
    \midrule
    \multirow{4}{*}{fr1/room}  & No Optimisation      &         0.267  &         0.227  &         0.583  \\
                               & Smoothing-Only       &         0.265  &         0.223  &         0.577  \\
                               & Total Variation      &         0.243  &         0.179  &         0.522  \\
                               & Normals + Occlusions & \textbf{0.226} & \textbf{0.147} & \textbf{0.488} \\
    \midrule
    \multirow{4}{*}{fr1/rpy}   & No Optimisation      &         0.327  &         0.434  &         0.781  \\
                               & Smoothing-Only       &         0.320  &         0.390  &         0.755  \\
                               & Total Variation      &         0.265  &         0.231  &         0.621  \\
                               & Normals + Occlusions & \textbf{0.255} & \textbf{0.212} & \textbf{0.614} \\
    \midrule
    \multirow{4}{*}{fr1/teddy} & No Optimisation      &         0.296  &         0.300  &         0.799  \\
                               & Smoothing-Only       &         0.295  &         0.296  &         0.792  \\
                               & Total Variation      & \textbf{0.290} & \textbf{0.276} & \textbf{0.771} \\
                               & Normals + Occlusions &         0.297  &         0.277  &         0.792  \\
    \midrule
    \multirow{4}{*}{fr1/xyz}   & No Optimisation      &         0.299  &         0.303  &         0.595  \\
                               & Smoothing-Only       &         0.296  &         0.298  &         0.590  \\
                               & Total Variation      &         0.255  &         0.212  &         0.493  \\
                               & Normals + Occlusions & \textbf{0.225} & \textbf{0.137} & \textbf{0.401} \\
    \bottomrule
\end{tabular}
\caption{Comparison of reconstruction errors on Freiburg 1 TUM RGB-D \cite{Sturm:etal:IROS2012} sequences showing the relative performance of different regularisation schemes. No Optimisation: results from taking the depth value with the maximum probability in the probability volume. Smoothing-Only: results from minimising the smoothed negative log probability density function without including a regularisation term. Total Variation: results from using the total variation of the depth as a regulariser. Normals + Occlusions: the pipeline as described in this paper.}
\label{tab:reg_ablation}
\vspace{2mm}\hrule
\end{table}

As discussed in the introduction, the best results (in terms of the accuracy of the final depth maps) seem to come from systems that take classic, photometric-based approaches and feed the results into a DNN for regularisation.
DeepTAM \cite{Zhou:etal:ECCV2018} is a state-of-the-art example of such a system.
We argue, however, that a probabilistic formulation is necessary for many applications of depth estimation in robotics and that it is important to investigate methods of fusing the output of learning-based systems into standard reconstruction pipelines that maintain this formulation.
We therefore do not expect or claim to be able to achieve more accurate depth reconstructions than those produced by DeepTAM.
Instead, we claim that we are able to improve the performance of standard SLAM systems by fusing in the outputs of a deep neural network while maintaining a probabilistic formulation.
For the sake of transparency, however, we have also run the DeepTAM mapping system with ground truth poses on the same sequences and have included the results in Table 1.

To show the benefit of our method of regularisation, we compare the performance of the full system against three other regularisation schemes: using no optimisation at all (taking the depth values that maximise the discrete probability distribution), optimising without any regularisation (this will allow for the smoothing of the depth maps based on the continuous PDF, but provide no regularisation), and regularising using the total variation.

For the total variation, we tuned the hyperparameters of our system for the best performance ($\lambda = 1.0 \cdot 10^2$ and a step size of 0.05).

The results are presented in Table \ref{tab:reg_ablation}. In all cases the best performance is achieved when using the surface normals and occlusion masks predicted by SharpNet.

Finally, to evaluate our method for warping probability volumes between keyframes, we compare our system against a version without warping where each keyframe is initialised only with the network output and does not receive any information from other keyframes.

The results are presented in Table \ref{tab:warp_ablation}.
Using our warping method improves the performance of the system in all cases.

\begin{table}[t]
\centering
\def\arraystretch{0.9}
\begin{tabular}{l l c c c}
    \toprule
    \textbf{Sequence} & \textbf{System} & \textbf{L1-rel} & \textbf{L2-rel} & \textbf{RMSE} \\
    \midrule
    \multirow{2}{*}{fr1/360}   & No Keyframe Warping &         0.202  &         0.157  &         0.575  \\
                               & Keyframe Warping    & \textbf{0.191} & \textbf{0.143} & \textbf{0.555} \\
    \midrule
    \multirow{2}{*}{fr1/desk}  & No Keyframe Warping &         0.316  &         0.236  &         0.474  \\
                               & Keyframe Warping    & \textbf{0.278} & \textbf{0.177} & \textbf{0.427} \\
    \midrule
    \multirow{2}{*}{fr1/desk2} & No Keyframe Warping &         0.283  &         0.195  &         0.480  \\
                               & Keyframe Warping    & \textbf{0.236} & \textbf{0.138} & \textbf{0.424} \\
    \midrule
    \multirow{2}{*}{fr1/floor} & No Keyframe Warping & \textbf{0.776} & \textbf{0.684} & \textbf{0.787} \\
                               & Keyframe Warping    &         0.785  &         0.691  &         0.796  \\
    \midrule
    \multirow{2}{*}{fr1/plant} & No Keyframe Warping &         0.420  &         0.490  &         0.845  \\
                               & Keyframe Warping    & \textbf{0.416} & \textbf{0.485} & \textbf{0.833} \\
    \midrule
    \multirow{2}{*}{fr1/room}  & No Keyframe Warping &         0.256  &         0.189  &         0.528  \\
                               & Keyframe Warping    & \textbf{0.226} & \textbf{0.147} & \textbf{0.488} \\
    \midrule
    \multirow{2}{*}{fr1/rpy}   & No Keyframe Warping &         0.297  &         0.263  &         0.654  \\
                               & Keyframe Warping    & \textbf{0.255} & \textbf{0.212} & \textbf{0.614} \\
    \midrule
    \multirow{2}{*}{fr1/teddy} & No Keyframe Warping &         0.302  &         0.286  & \textbf{0.791} \\
                               & Keyframe Warping    & \textbf{0.297} & \textbf{0.277} &         0.792  \\
    \midrule
    \multirow{2}{*}{fr1/xyz}   & No Keyframe Warping &         0.315  &         0.247  &         0.521  \\
                               & Keyframe Warping    & \textbf{0.225} & \textbf{0.137} & \textbf{0.401} \\
    \bottomrule
\end{tabular}
\caption{Comparison of reconstruction errors on Freiburg 1 TUM RGB-D \cite{Sturm:etal:IROS2012} sequences showing the performance gain from using our method to warp keyframe probability volumes.}
\label{tab:warp_ablation}
\vspace{2mm}\hrule
\end{table}

\section{CONCLUSION}

We have presented a method for fusing learned monocular depth priors into a standard pipeline for 3D reconstruction.
By training a DNN to predict nonparametric probability distributions, we allow the network to express uncertainty and make multi-hypothesis depth predictions.

Through a series of experiments, we demonstrated that by fusing the discrete probability volume predicted by the network with a probability volume computed from the photometric error, we often achieve better performance than either on its own.
Further experiments showed the value of our regularisation scheme and warping method.

\bibliographystyle{IEEEtran}
\bibliography{IEEEabrv,robotvision}

\end{document}